\documentclass{article}

\usepackage[final]{nips_2017}

\usepackage[utf8]{inputenc} 
\usepackage[T1]{fontenc}    
\usepackage{hyperref}       
\usepackage{url}            
\usepackage{booktabs}       
\usepackage{amsfonts}       
\usepackage{nicefrac}       
\usepackage{microtype}      

\title{``I know it when I see it''. Visualization and Intuitive Interpretability}

\author{
  Fabian Offert\\
  Machine Learning and the Arts Research Group\\
  Media Arts and Technology Program\\
  University of California, Santa Barbara\\
  \texttt{offert@ucsb.edu} \\
}

\begin{document}

\maketitle

\begin{abstract}
  Most research on the interpretability of machine learning systems focuses on the development of a more rigorous notion of interpretability. I suggest that a better understanding of the deficiencies of the intuitive notion of interpretability is needed as well. I show that visualization enables but also impedes intuitive interpretability, as it presupposes two levels of technical pre-interpretation: dimensionality reduction and regularization. Furthermore, I argue that the use of positive concepts to emulate the distributed semantic structure of machine learning models introduces a significant human bias into the model. As a consequence, I suggest that, if intuitive interpretability is needed, singular representations of internal model states should be avoided.
\end{abstract}

\section{Re-investigating Intuitive
Interpretability}\label{re-investigating-intuitive-interpretability}

Philip Agre has argued that ``technology at present is covert
philosophy'' (Agre, 1997). While the scope of this claim is certainly
debatable, in the quest for interpretable machine learning models,
certain philosophical issues are evident, above all the fact that
interpretability itself is an intuitive notion. As Wolfgang Iser
remarks: ``For a long time, interpretation was taken for an activity
that did not seem to require analysis of its own procedures. There was a
tacit assumption that it came naturally, not least because human beings
live by constantly interpreting.'' (Iser, 2000)

Nevertheless, Kim and Doshi-Velez (2017) and many others have shown that
interpretability can be transformed into a more rigorous notion. While
most investigations into the interpretability of machine learning models
thus focus on the further development of this rigorous notion of
interpretability, I suggest that a re-investigation of the intuitive
notion of interpretability can help to better understand the limits of
interpretability in general.

When we talk about intuitive interpretability in the context of machine
learning, we assume a Cartesian concept of intuition that posits
intuitive concepts as rational concepts, and thus intuition as an
adequate measure of reality. The statement ``I know it when I see it'',
which is often employed to illustrate this concept, indicates the
dependency of such intuition on visualization. To intuitively understand
a machine learning model, we need to visualize it, make it accessible to
the senses. This process, however, is not as straightforward as it
seems. I argue that specifically for machine learning models,
visualization -- and thus intuitive interpretation -- necessarily
implies two levels of pre-interpretation.

\section{Intuitive Interpretability Depends on Dimensionality
Reduction}\label{intuitive-interpretability-depends-on-dimensionality-reduction}

Machine learning models operate in high-dimensional vector spaces.
High-dimensional vector spaces are geometrically counter-intuitive.
While low-dimensional vector spaces can always be intuitively correlated
with our physical reality, with the existence of objects in space and
time, high-dimensional vector spaces have no intuitive equivalent in the
real world. Beyond this general inaccessibility, however,
high-dimensional vector spaces also specifically impede interpretation,
as distances between data points have a tendency to lose their
\emph{meaning} (Beyer et al., 1999) -- this is commonly known as the
``curse of dimensionality''. Making a high-dimensional vector space
intuitively interpretable thus requires its mathematical
pre-interpretation, its representation in human terms, i.e.~usually in
no more than three dimensions.

While there is certainly a quantifiable limit to the damage
dimensionality reduction can inflict (Johnson and Lindenstrauss, 1984)
it is nevertheless important to acknowledge the reason for the
inevitability of this mathematical pre-interpretation. Internal states
of machine learning models are non-concepts, concepts that have no
intuitive equivalent in the real world and that can only be represented
\emph{in terms of what they are not}. This notion of the non-concept
will guide our further investigation of intuitive interpretability.

\section{Intuitive Interpretability Depends on
Regularization}\label{intuitive-interpretability-depends-on-regularization}

Artificial neural networks trained on image data are notoriously opaque.
Particularly for deep convolutional neural networks (Krizhevsky et al.,
2012), it is very hard to infer from the training dataset and the final
weights of the fully trained neural network how exactly the network
makes its decisions. Many different approaches to this problem have been
suggested, most prominently two types of feature visualization:
activation maximization (Zeiler and Fergus, 2014, Simonyan et al.
(2014), Mahendran and Vedaldi (2015), Mahendran and Vedaldi (2016),
Nguyen, Dosovitskiy, et al. (2016), Nguyen, Yosinski, et al. (2016)) and
saliency maps, a technique also called attribution (Olah et al., 2017,
Simonyan et al. (2014), Zeiler and Fergus (2014)). We will focus here on
activation maximization.

Naive optimizations of an image to maximally activate a specific
``neuron'', i.e.~the part of an artificial neural network that encodes a
specific feature, often result in noise and ``nonsensical high-frequency
patterns'' (Olah et al., 2017) -- patterns that are without meaning and
are thus, again, inaccessible to an intuitive interpretation. The
regularization of this optimization is thus another pre-interpetation
that is necessary to establish intuitive interpretability. The goal of
activation maximizations is thus to generate ``natural'' pre-images
(Mahendran and Vedaldi, 2015, Mahendran and Vedaldi (2016)) -- images
that are visual representations of intermediate stages in the neural
network, expressed \emph{in terms of} a set of natural images. This
regularization is achieved by introducing natural image priors into the
objective function.

\section{Non-Concepts and the Place of Semantic
Information}\label{non-concepts-and-the-place-of-semantic-information}

The natural pre-images of activation maximization usually consist of an
arbitrary ``mix'' of different representations. This mix of
representations can either be a set of images that each show a different
aspect of the activation maximization, or a ``blend'' of different
images, i.e.~a single image that maximizes different aspects of the
neuron. Most recently, such ``multifaceted'' representations have been
improved significantly (Nguyen, Dosovitskiy, et al., 2016, Nguyen,
Yosinski, et al. (2016)) through the automatic generation of natural
image priors with the help of an additional, generative adversarial
neural network. Other approaches have used techniques from style
transfer to likewise increase the ``diversity'' (Olah et al., 2017) of
the visualization.

However, as Olah et al. (2017) observe, many of the resulting images are
``strange mixtures of ideas'' suggesting that single neurons are not
necessarily the right semantic units for understanding neural nets." In
fact, as Szegedy et al. (2013) showed, looking for meaningful features
does not necessarily lead to more meaningful visualizations than looking
for any features, i.e.~for arbitrary activation maximizations. This is
also the reason for the effectiveness of many adversarial strategies (Su
et al., 2017, Papernot et al. (2017), Kurakin et al. (2016), Goodfellow
et al. (2014)).

In other words, not only is the representation of non-concepts mediated
twice, by means of dimensionality reduction and regularization, it is
also questionable if non-concepts can be approximated at all in human
terms. Szegedy et al. (2013) suggest that the entire space of
activations, rather than the individual units contain most of the the
semantic information.\footnote{While we will not develop this idea
  further within this limited context, it is worth noting that the
  findings in Szegedy et al. (2013) in relation to the notion of
  non-concepts mirror long-standing discussions in the humanities on
  meaning and interpretation, with the most prominent concept being
  Jacques Derrida's notion of différance (Derrida, 1982).} As also
pointed out in Szegedy et al. (2013), a ``similar but even stronger
conclusion'' was reached for word embedding models, and in fact the
concept of a distributed semantic structure becomes even more obvious
when we look at the text and not the image domain.

Word embedding models (Mikolov et al., 2013) employ shallow artificial
neural networks to construct high-dimensional vector spaces that not
only reflect syntactic but also semantic properties of the source
corpus. Most prominently, word embedding models are able to solve
analogy queries, like ``what is to woman what king is to man''. This is
achieved by not extending, but reducing the dimensionality of the vector
space in relation to the number of n-grams in the source corpus.
Accordingly, no vector represents just a single n-gram. Instead, the
totality of vectors represents the totality of the semantic structure of
the source corpus. This distributed semantic structure, however, has
peculiar consequences. The solution to an analogy query is given by the
model not as a definite answer, but as a hierarchy of answers. Why?
Simply because there are no ``intermediate'' words. If the best possible
analogy is a (new) data point right in between two (existing) data
points representing n-grams in the source corpus vocabulary, the best
possible solution to the analogy query is neither of them, but it still
can only be described \emph{in terms of} them. Even if the input
vocabulary consisted of all words in the English language, the solution
to the analogy query could still be a data point that is ``in between
everything'' but has no equivalent in the real world -- a non-concept.
Every computational solution to an analogy task is thus, ironically,
itself an analogy.

\section{Non-Concepts as a Critical Technical Practice: Revealing Human
Bias}\label{non-concepts-as-a-critical-technical-practice-revealing-human-bias}

A demonstration of this dilemma of non-concepts, and an example for a
critical technical practice based on it, is ``Image Synthesis from
Yahoo's open\_nsfw''(Goh, 2016), a project by Gabriel Goh. Using the
technique developed in (Nguyen, Dosovitskiy, et al., 2016) Goh produces
images that maximally activate certain neurons of a classifier network
called ``open\_nsfw'', which was created by Yahoo to distinguish
workplace-safe (``SFW'') from ``not-safe-for-work'' (``NSFW'') imagery:
a literal mathematical model of ``I know it when I see it''. By
generating sets of images ranging from most to least pornographic, Goh
produces some interesting insights into Yahoo's specific interpretation
of ``nsfw'', and the essence of the concept of pornography. Most
interesting, however, are the ``least pornographic'' images. What really
\emph{is} the ``opposite'' of pornography? Fully clothed people?
Non-pornography, again, is a non-concept which has never been defined
but through its negation.

Except in this particular case it hasn't. Goh notes that the least
pornographic images ``all have a distinct pastoral quality -- depictions
of hills, streams and generally pleasant scenery'' and concludes that,
most likely, this is the result of providing negative examples during
training . Apparently, the non-concept of non-pornography was made into
a positive concept -- pastoral landscapes -- to improve the training of
the model. This, of course, becomes particularly problematic if, as it
is the case with open\_nsfw, a fully trained model is provided without
access to the training data. While the Github page for open\_nsfw
acknowledges that the ``definition of NSFW is subjective and
contextual'', what is at stake here is exactly the opposite: the fact
that ``SFW'' is subjective and contextual, and that regardless a very
specific notion of ``SFW'' was built into the model. More generally
speaking, the approximation of non-concepts with positive concepts
necesarily introduces a significant human -- aesthetic -- bias into the
equation.

\section{Conclusion}\label{conclusion}

Goh's project serves to show the non-conceptual structure of machine
learning models and the problems this structure creates for intuitive
interpretability. While one possible way to address these problems (if
intuitive interpretability is needed) is to avoid strategies that
present singular images as representations of internal model states
altogether, and instead switch to multitudes of images, there is no
general solution to the problem of finding human-readable
representations of non-concepts. Research in interpretability thus has
to take the non-conceptual structure of machine learning models into
account. To make the notion of interpretability more rigorous we have to
first identify where it might still be impaired by intuitive
considerations: we have to consider it precisely in terms of what it is
not.

\section*{References}

\small

Agre, P.E., 1997. Computation and Human Experience. Cambridge University
Press.

Beyer, K., Goldstein, J., Ramakrishnan, R., Shaft, U., 1999. When is
``nearest neighbor'' meaningful?, in: International Conference on
Database Theory. Springer, pp. 217--235.

Derrida, J., 1982. Différance, in: Margins of Philosophy. University of
Chicago Press.

Goh, G., 2016. Image synthesis from Yahoo's open\_nsfw. Blog: Gabriel Goh.

Goodfellow, I.J., Shlens, J., Szegedy, C., 2014. Explaining and
harnessing adversarial examples. arXiv preprint arXiv:1412.6572.

Iser, W., 2000. The Range of Interpretation. Columbia University Press,
New York, NY.

Johnson, W.B., Lindenstrauss, J., 1984. Extensions of Lipschitz mappings
into a Hilbert space. Contemporary Mathematics 26, 1.

Kim, B., Doshi-Velez, F., 2017. Towards a rigorous science of
interpretable machine learning. arXiv preprint arXiv:1702.08608.

Krizhevsky, A., Sutskever, I., Hinton, G.E., 2012. Imagenet
classification with deep convolutional neural networks, in: Advances in
Neural Information Processing Systems. pp. 1097--1105.

Kurakin, A., Goodfellow, I., Bengio, S., 2016. Adversarial examples in
the physical world. arXiv preprint arXiv:1607.02533.

Mahendran, A., Vedaldi, A., 2016. Visualizing deep convolutional neural
networks using natural pre-images. International Journal of Computer
Vision 120, 233--255.

Mahendran, A., Vedaldi, A., 2015. Understanding deep image
representations by inverting them, in: Proceedings of the IEEE
Conference on Computer Vision and Pattern Recognition. pp. 5188--5196.

Mikolov, T., Sutskever, I., Chen, K., Corrado, G.S., Dean, J., 2013.
Distributed representations of words and phrases and their
compositionality, in: Advances in Neural Information Processing Systems.
pp. 3111--3119.

Mordvintsev, A., Olah, C., Mike, T., 2015. Inceptionism: Going deeper
into neural networks. Google Research Blog.

Nguyen, A., Dosovitskiy, A., Yosinski, J., Brox, T., Clune, J., 2016.
Synthesizing the preferred inputs for neurons in neural networks via
deep generator networks, in: Advances in Neural Information Processing
Systems. pp. 3387--3395.

Nguyen, A., Yosinski, J., Clune, J., 2016. Multifaceted feature
visualization: Uncovering the different types of features learned by
each neuron in deep neural networks. arXiv preprint arXiv:1602.03616.

Olah, C., Mordvintsev, A., Schubert, L., 2017. Feature visualization.
Distill.

Papernot, N., McDaniel, P., Goodfellow, I., Jha, S., Celik, Z.B., Swami,
A., 2017. Practical black-box attacks against machine learning, in:
Proceedings of the 2017 ACM Asia Conference on Computer and
Communications Security. ACM, pp. 506--519.

Simonyan, K., Vedaldi, A., Zisserman, A., 2014. Deep inside
convolutional networks: Visualising image classification models and
saliency maps. arXiv preprint arXiv:1312.6034.

Su, J., Vargas, D.V., Kouichi, S., 2017. One pixel attack for fooling
deep neural networks. arXiv preprint arXiv:1710.08864.

Szegedy, C., Zaremba, W., Sutskever, I., Bruna, J., Erhan, D.,
Goodfellow, I., Fergus, R., 2013. Intriguing properties of neural
networks. arXiv preprint arXiv:1312.6199.

Zeiler, M.D., Fergus, R., 2014. Visualizing and understanding
convolutional networks, in: European Conference on Computer Vision.
Springer, pp. 818--833.

\end{document}